\RequirePackage{amsmath}
\documentclass[runningheads]{llncs}
\usepackage{ulem}
\usepackage{amsmath}
\usepackage{graphicx}
\usepackage{siunitx}
\usepackage{cite}
\usepackage{makecell}
\DeclareMathOperator*{\argmax}{argmax}
\DeclareMathOperator*{\argmin}{argmin}
\usepackage{boldline,multirow}
\usepackage{color}
\usepackage{bbm}
\usepackage{soul} 
\usepackage[colorlinks=true,citecolor=blue,urlcolor=black]{hyperref}

\usepackage{bbding}
\newcommand{\tabincell}[2]{\begin{tabular}{@{}#1@{}}#2\end{tabular}}
\usepackage{textcomp}
\usepackage{gensymb}
\usepackage{bbding}

\begin{document}
%

\title{Estimating Model Performance under Domain Shifts with Class-Specific Confidence Scores}
\titlerunning{Estimating Model Performance under Domain Shifts}
%

\newif\ifreview

\ifreview
\author{Paper ID 439}
\institute{--}
\else
\author{Zeju Li\inst{1} (\Envelope), Konstantinos Kamnitsas\inst{2,1,3}, Mobarakol Islam\inst{1}, Chen Chen\inst{1} and Ben Glocker\inst{1}}
\institute{BioMedIA Group, Department of Computing, Imperial College London, UK \and Institute of Biomedical Engineering, Department of Engineering Science, University of Oxford, Oxford, UK \and School of Computer Science, University of Birmingham, Birmingham, UK \email{zeju.li18@imperial.ac.uk} }
\authorrunning{Z. Li, K. Kamnitsas, M. Islam, C. Chen and B. Glocker}
\fi


%
\maketitle              
\begin{abstract}
Machine learning models are typically deployed in a test setting that differs from the training setting, potentially leading to decreased model performance because of domain shift. If we could estimate the performance that a pre-trained model would achieve on data from a specific deployment setting, for example a certain clinic, we could judge whether the model could safely be deployed or if its performance degrades unacceptably on the specific data. Existing approaches estimate this based on the confidence of predictions made on unlabeled test data from the deployment's domain. We find existing methods struggle with data that present class imbalance, because the methods used to calibrate confidence do not account for bias induced by class imbalance, consequently failing to estimate class-wise accuracy. Here, we introduce class-wise calibration within the framework of performance estimation for imbalanced datasets. Specifically, we derive class-specific modifications of state-of-the-art confidence-based model evaluation methods including temperature scaling (TS), difference of confidences (DoC), and average thresholded confidence (ATC). We also extend the methods to estimate Dice similarity coefficient (DSC) in image segmentation. We conduct experiments on four tasks and find the proposed modifications consistently improve the estimation accuracy for imbalanced datasets. Our methods improve accuracy estimation by 18\% in classification under natural domain shifts, and double the estimation accuracy on segmentation tasks, when compared with prior methods\footnote{We provide the source code for our experiments at \url{https://github.com/ZerojumpLine/ModelEvaluationUnderClassImbalance}}.

\end{abstract}
\section{Introduction}

Based on the independently and identically distributed (IID) assumption, one can estimate the model performance with labeled validation data which is collected from the same source domain as the training data. However, in many real-world scenarios where test data is typically sampled from a different distribution, the IID assumption does not hold anymore and the model's performance tends to degrade because of domain shifts~\cite{ben2010theory}. It is difficult for the practitioner to assess the model reliability on the target domain as labeled data for a new test domain is usually not available. It would be of great practical value to develop a tool to estimate the performance of a trained model on an unseen test domain without access to ground truth~\cite{eche2021toward}. For example, for a skin lesion classification model which is trained with training data from the source domain, we can decide whether to deploy it to a target domain based on the predicted accuracy of a few unlabeled test images. In this study we aim to estimate the model performance by only making use of unlabeled test data from the target domain. 

Many methods have been proposed to tackle this problem and they can be categorized into three groups. The first line of research is based on the calculation of a distance metric between the training and test datasets~\cite{chen2021mandoline, deng2021labels}. The model performance can be obtained by re-weighting the validation accuracy~\cite{chen2021mandoline} or training a regression model~\cite{deng2021labels}. The second type of methods is based on reverse testing which trains a new reverse classifier using test images and predicted labels as pseudo ground truth~\cite{fan2006reverse, valindria2017reverse}. The model performance is then determined by the prediction accuracy of the reverse classifier on the validation data. The third group of methods is based on model confidence scores~\cite{guo2017calibration, guillory2021predicting, garg2022leveraging}. The accuracy of a trained model can be estimated with the Average Confidence (AC) which is the mean value of the maximum softmax confidence on the target data. Previous studies have shown that confidence-based methods show better performance than the first two methods~\cite{elsahar2019annotate, garg2022leveraging} and are efficient as they do not need to train additional models. We focus on confidence-based methods in this study.

\begin{figure}[t]
\centering
\includegraphics[width=0.9\textwidth]{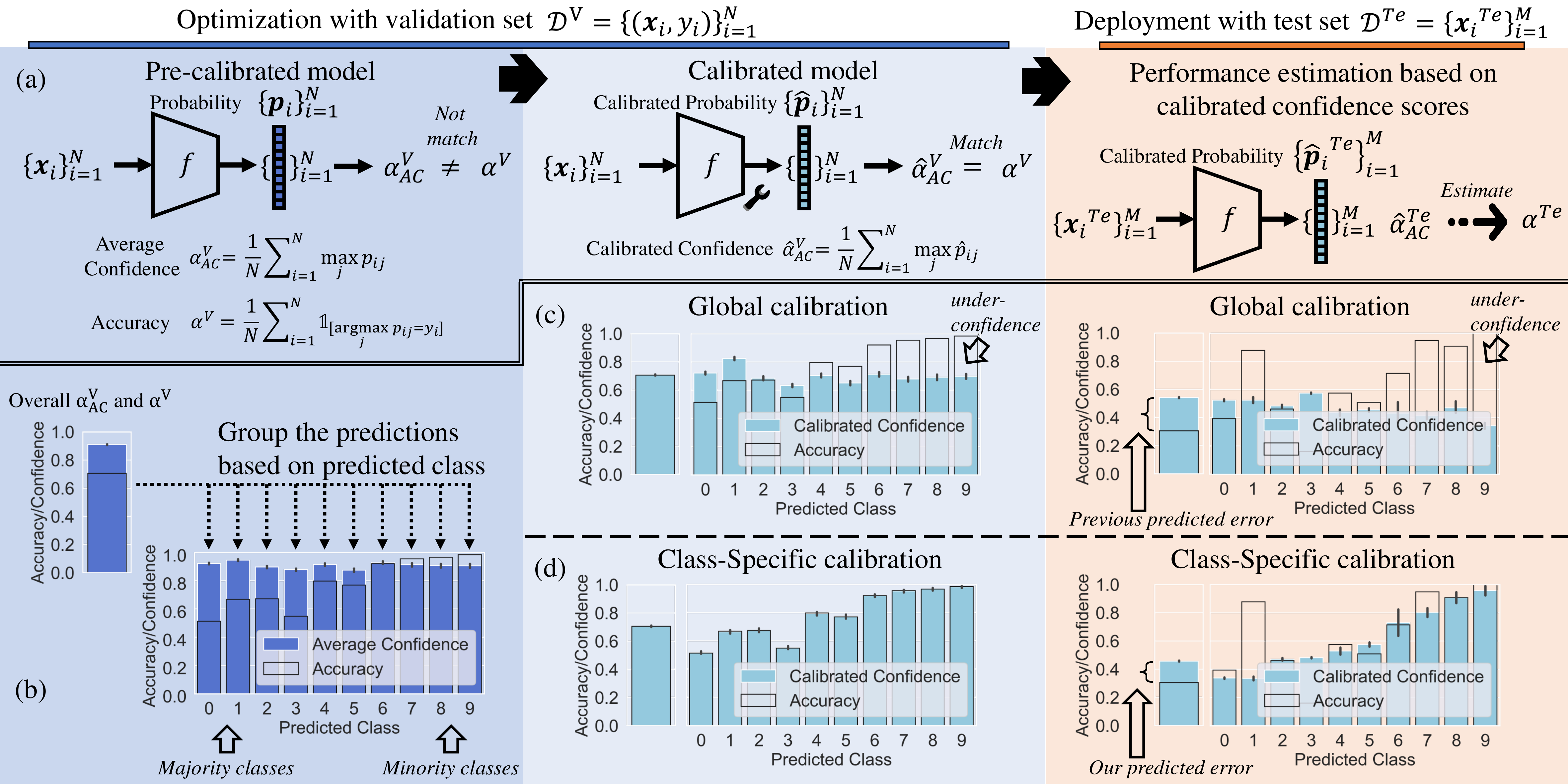}

\caption{Effect of class imbalance on confidence-based model evaluation methods. (a) The pipeline for model evaluation~\cite{guillory2021predicting,garg2022leveraging}, which estimates the accuracy under domain shifts with Average Confidence (AC) after obtaining calibrated probabilities $\hat{\boldsymbol{p}_i}$. (b) The confidence scores are biased towards the majority classes. (c) Prior methods reduce all probabilities in the same manner to match overall AC with the overall accuracy, but fail to calibrate class-wise probability. (d) Our proposed method with class-specific calibration parameters successfully reduces the bias and significantly increases the model estimation accuracy. The shown example is for imbalanced CIFAR-10.} \label{fig1}
\end{figure}

The prediction results by AC are likely to be higher than the real accuracy as modern neural networks are known to be over-confident~\cite{guo2017calibration}. A few methods, such as Temperature Scaling (TS)~\cite{guo2017calibration}, Difference of Confidences (DoC)~\cite{guillory2021predicting}, and Average Thresholded Confidence (ATC)~\cite{garg2022leveraging}, have been proposed to calibrate the probability of a trained model with validation data with the expectation to generalize well to unseen test data, as depicted in Fig.~\ref{fig1}(a). We argue those methods do not generalize well to imbalanced datasets, which are common in medical image classification and segmentation. Specifically, previous work found that a model trained with imbalanced data is prone to predict a test sample from a minority class as a majority class, but not vice versa~\cite{li2020analyzing}. As a result, such models tend to have high classification accuracy but are under-confident on samples predicted as the minority classes, with opposite behaviour for the majority classes, as illustrated in Fig.~\ref{fig1}(b). Current post-training calibration methods adopt the same parameter to adjust the probability, which would increase the estimation error for the minority class samples and thus, does not generalize well, as shown in Fig.~\ref{fig1}(c). We introduce class-specific parameters into the performance estimation framework using adaptive optimization, alleviating the confidence bias of neural networks against minor classes, as demonstrated in Fig.~\ref{fig1}(d). It should be noted that our method differs from other TS methods with class-wise parameters such as vector scaling (VS)~\cite{guo2017calibration}, class-distribution-aware TS~\cite{islam2021class} and normalized calibration (NORCAL)~\cite{pan2021model} as 1) our parameters are optimized so that class-specific confidence matches class-specific accuracy, whereas their parameters are optimized such that overall-confidence matches overall-accuracy; 2) we determine the calibrated class based on prediction instead of ground truth, therefore the model predictions would not be affected by the calibration process; 3) we apply the proposed methods to model evaluation; 4) we also consider other calibration methods than TS.

This study sheds new light on the problem of model performance estimation under domain shifts in the absence of ground truth with the following contributions: 1) We find existing solutions encounter difficulties in real-world datasets as they do not consider the confidence bias caused by class imbalance; 2) We propose class-specific variants for three state-of-the-art approaches including TS, DoC and ATC to obtain class-wise calibration; 3) We extend the confidence-based model performance framework to segmentation and prediction of Dice similarity coefficient (DSC); 4) Extensive experiments performed on four tasks with a total of 310 test settings show that the proposed methods can consistently and significantly improve the estimation accuracy compared to prior methods. Our methods can reduce the estimation error by 18\% for skin lesion classification under natural domain shifts and double the estimation accuracy in segmentation.

\section{Method}
\label{sec:method}

\noindent\textbf{Problem setup.} We consider the problem of $c$-class image classification. We train a classification model $f$ that maps the image space $\mathcal{X} \subset \mathbbm{R}^d$ to the label space $\mathcal{Y} = \{1,2, \ldots ,c\}$. We assume that two sets of labeled images: $\mathcal{D}^{Tr} = \{ (\boldsymbol{x}_i, y_i) \}_{i = 1}^I$ and $\mathcal{D}^V = \{ (\boldsymbol{x}_i, y_i) \}_{i = 1}^N$ drawn from the \emph{same} source domain are available for model training and validation. After being trained, the model is applied to a target domain $\mathcal{D}^{Te}$ whose distribution is \emph{different} from the source domain, which is typical in the real-world applications. We aim to estimate its accuracy $\alpha^{Te}$ on the target domain to support decision making when there is only a set of test images $\mathcal{D}^{Te} = \{ \boldsymbol{x}^{Te}_i \}_{i = 1}^M$ available \emph{without any ground-truth}.

\begin{figure}[t]
\centering
\includegraphics[width=0.85\textwidth]{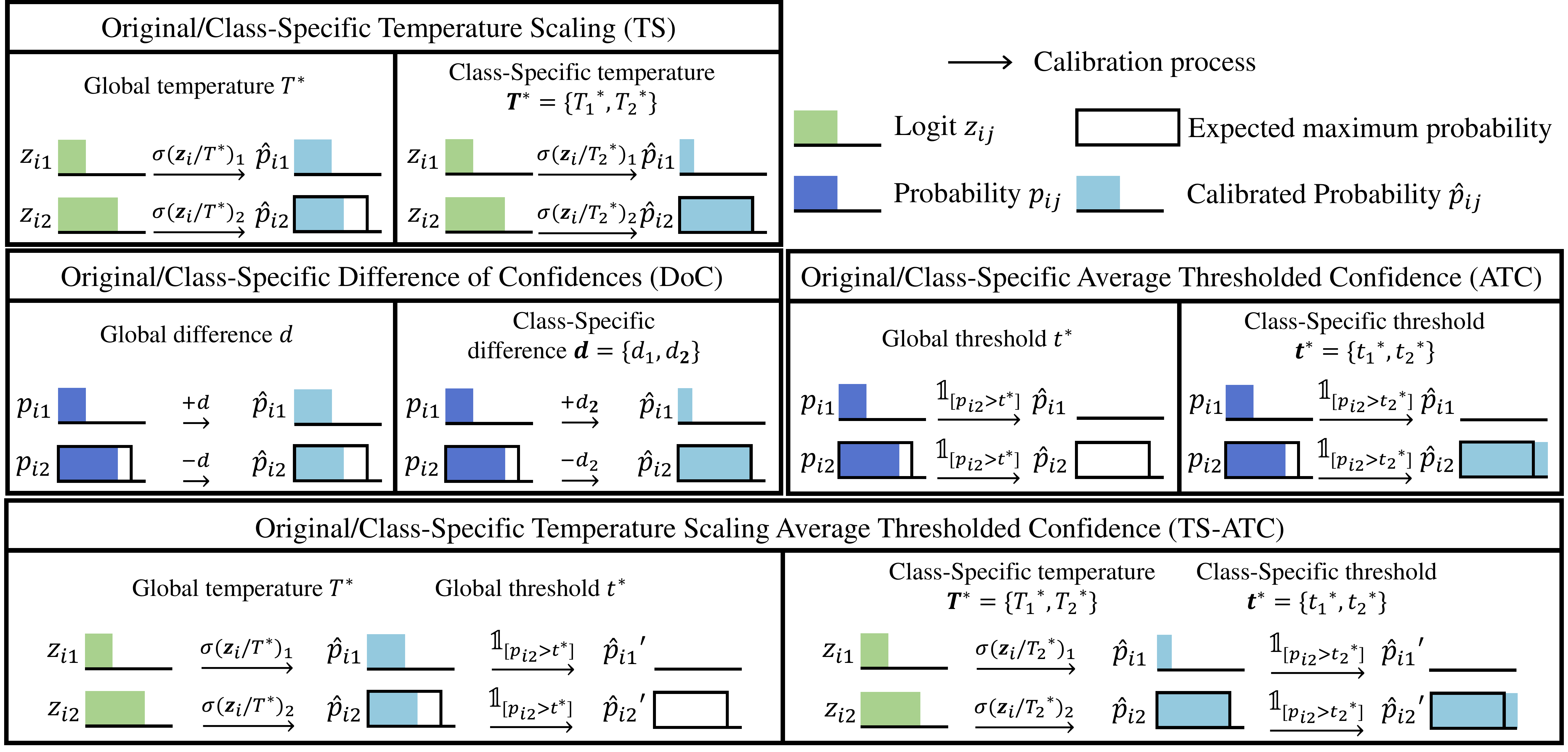}

\caption{Illustration of proposed Class-Specific modifications for four existing model evaluation methods. We show the calibration process of an under-confident prediction made for sample from minority class $c\!=\!2$. Prior calibration methods use a global parameter for all classes, which leads to sub-optimal calibration and therefore bias for the minority class. The proposed variants adapt separate parameters per class, enabling improved, class-wise calibration.} \label{fig2}
\end{figure}

\noindent\textbf{A unified test accuracy framework based on model calibration.} To achieve this goal, we first provide a unified test accuracy estimation framework that is compatible with existing confidence-based methods. Let $\boldsymbol{p}_i$ be the predicted probability for input $\boldsymbol{x}_i$, which is obtained by applying the softmax function to the network output logit: $\boldsymbol{z}_i = f(\boldsymbol{x}_i)$. The predicted probability for the $j$'th class is computed as $p_{ij} = \sigma(\boldsymbol{z}_i)_j = \frac{\mathrm{e}^{z_{ij}}}{\sum_{j=1}^{c}\mathrm{e}^{z_{ij}}}$, and the predicted class $y^{\prime}_i$ is associated with the one of the largest probability: $y^{\prime}_i= \argmax_j p_{ij}$. The test accuracy on the test set $\alpha^{Te}$ can be estimated via Average Confidence (AC)~\cite{guo2017calibration}: $\alpha_{AC}^{Te} = \frac{1}{N}\sum_{i=1}^{N}\max_j p_{ij}^{Te}$, which is an average of the probability of the predicted class $j$: $\max_j p_{ij}^{Te}$ for every test image $\boldsymbol{x}_{i}^{Te}$. Yet, directly computing over the softmax output $p_{ij}^{Te}$ for estimation can be problematic as the predicted probability cannot reflect the real accuracy with modern neural networks~\cite{guo2017calibration}. It is essential to use calibrated probability $\hat{p}_{ij}^{Te}$ instead to estimate model accuracy with $\hat{\alpha}_{AC}^{Te} = \frac{1}{N}\sum_{i=1}^{N}\max_j \hat p_{ij}^{Te}$. Ideally, the calibrated probability $\hat p_{ij}^{Te}$ is supposed to perfectly reflect the expectation of the real accuracy. To adjust $\hat{p}_{ij}$, it is a common practice to use validation accuracy, which is defined as $\alpha^V=\frac{1}{N}\sum_{i=1}^{N}\mathbbm{1}_{[y^{\prime}_i=y_i]}$\footnote{ $\mathbbm{1}$ is an indicator function which is equal to 1 when the underlying criterion satisfies and $0$ otherwise.}, as an surrogate objective to find the solution as there is no ground truth labels for test data~\cite{elsahar2019annotate, guillory2021predicting, garg2022leveraging}. On top of this pipeline which is also illustrated in Fig.~\ref{fig1}(a), we will introduce state-of-the-art calibration methods including TS, DoC, and ATC and our proposed class-specific modifications of them (cf. Fig.~\ref{fig2}), alleviating the calibration bias caused by the common class imbalance issue in medical imaging mentioned earlier.

\noindent\textbf{Class-Specific Temperature Scaling.} The TS method~\cite{guo2017calibration} rescales the logits using temperature parameter $T$. Calibrated probability is obtained with $\hat{p}_{ij} = \sigma(\boldsymbol{z}_i/T)_j$. To find $T$ such that the estimated accuracy will approximate the real accuracy, we obtain optimal $T^*$ using $\mathcal{D}^V$ with:

\begin{equation}
    T^{*} = \argmin_T\Big|\frac{1}{N}\sum_{i=1}^{N}\max_j \sigma(\frac{\boldsymbol{z}_i}{T})_j-\alpha^V\Big|.
    \label{eq:TS}
\end{equation}

To enable class-specific calibration, we extend temperature to vector $\boldsymbol{T}$ with $c$ elements, where $j$'th element $T_j$ corresponds to the $j$'th class. The calibrated probability is then written as $\hat{p}_{ij} = \sigma(\boldsymbol{z}_i/T_{y^{\prime}_i}^*)_j$, where $T_j^*$ is optimized with:

\begin{equation}
    T_j^* = \argmin_{T_j}\Big|\frac{1}{N_j}\sum_{i=1}^{N}\mathbbm{1}_{ij}\max_j \sigma(\frac{\boldsymbol{z}_i}{T_j})_j-\alpha^V_j\Big|,
    \label{eq:CSTS}
\end{equation}

where $\mathbbm{1}_{ij} = \mathbbm{1}_{[y^{\prime}_i=j]}$ is 1 if the predicted most probable class is $j$ and $0$ otherwise, $N_j=\sum_{i=1}^{N}\mathbbm{1}_{ij}$ and $\alpha^V_j$ is the accuracy for class $j$ on the validation set which is calculated as $\alpha^V_j=\frac{1}{N_j}\sum_{i=1}^{N}\mathbbm{1}_{ij}\mathbbm{1}_{[y^{\prime}_i=y_i]}$. This way we ensure optimized $\boldsymbol{T}^*$ leads to $\hat{\boldsymbol{p}}_i$ matching the class-wise accuracy.

\noindent\textbf{Class-Specific Difference of Confidences.} The DoC method~\cite{guillory2021predicting} adjusts $p_i$ by subtracting the difference $d = \alpha^V_{AC} - \alpha^V$ between the real accuracy $\alpha^V$ and $\alpha^V_{AC}$ from the maximum probability:

\begin{equation}
\hat{p}_{ij}=\left\{
\begin{array}{lcl}
p_{ij} - d         &      & \text{if} \quad y^{\prime}_i=j,\\
p_{ij} + \frac{d}{c-1}       &      & \text{otherwise}.
\end{array} \right.
\label{eq:DoC}
\end{equation}

The two parts of Eq.~\ref{eq:DoC} ensure that the class-wise re-calibrated probabilities remain normalized. We propose a class-specific DoC method, where we define the difference vector $\boldsymbol{d}$ with $c$ elements, with its $j$'th element $d_j$ corresponding to the $j$'th class's difference. The class-wise calibrated probability $\hat{p}_{ij}$ is then obtained via:

\begin{equation}
\hat{p}_{ij}=\left\{
\begin{array}{lcl}
p_{ij} - d_{j}         &      & \text{if} \quad y^{\prime}_i=j,\\
p_{ij} + \frac{d_{y^{\prime}}}{c-1}       &      & \text{otherwise},
\end{array} \right.
\label{eq:CSDoC}
\end{equation}where $d_j = \frac{1}{N_j} \sum_{i=1}^{N} \mathbbm{1}_{ij}\max_j p_{ij} - \alpha^V_j$ is the difference between the average predicted probability for samples predicted that they belong to class $j$ and $\alpha^V_j$. In this way, the probabilities of samples which are predicted as the minority classes would not be affected by the majority classes and thus be less biased.

\noindent\textbf{Class-Specific Average Thresholded Confidence.} The original ATC~\cite{garg2022leveraging} proposes predicting the accuracy as the portion of samples with $\boldsymbol{p}_i$ that is higher than a learned threshold $t$. $t$ is obtained via:
\begin{equation}
    t^{*} = \argmin_t\Big|\frac{1}{N}\sum_{i=1}^{N}\mathbbm{1}_{[\max_j p_{ij}>t]} - \alpha^V\Big|.
    \label{eq:ATC}
\end{equation}
We can then use $t^{*}$ to compute $\hat{p}_{ij} = \mathbbm{1}_{[p_{ij}>t^*]}$ for each test sample to calculate the estimated test accuracy, $\hat{\alpha}_{AC}^{Te}$, defined previously. We should note that for ATC, $\hat{p}_{ij}$ is no longer a calibrated probability but a discrete number $\in \{0, 1\}$. For class-specific calibration, we propose learning a set of thresholds given by vector $\boldsymbol{t}$ with $c$ elements, one per class, where $j$'th element is obtained by:

\begin{equation}
    t_j^* = \argmin_{t_j}\Big|\frac{1}{N_j}\sum_{i=1}^{N}\mathbbm{1}_{ij}\mathbbm{1}_{[max_k p_{ik}>t_j]} - \alpha^V_j\Big|.
    \label{eq:CSATC}
\end{equation}

We can then calibrate probabilities as $\hat{p}_{ij} = \mathbbm{1}_{[p_{ij}>t_{y^{\prime}_i}^*]}$ based on predicted class $y^{\prime}_i$. Hence the probabilities of samples that are predicted as different classes are discretized using different thresholds $t_{y^{\prime}_i}^*$ to compensate the bias incurred by class imbalance. ATC can be combined with TS, by applying it to $\boldsymbol{\hat{p}}_{i}$ which is calibrated by TS, and we denote this method as TS-ATC. By combining our class-specific modifications of TS and ATC, we similarly obtain the class-specific variant CS TS-ATC (cf. Fig.~\ref{fig2}).

\noindent\textbf{Extension to Segmentation.} We propose extending confidence-based model evaluation methods to segmentation by predicting the DSC that the model would achieve on target data, instead of the accuracy metric that is more appropriate for classification. For this purpose, we slightly modify notation and write that the validation dataset consists of $Z$ segmentation cases as $\mathcal{D}^V = \{ \mathcal{S}_z \}_{z = 1}^Z$, with $\mathcal{S}_z$ all the training pairs for the case $z$ consisting of totally $n$ pixels, defined as $\mathcal{S}_z = \{(\boldsymbol{x}_{zi}, y_{zi}) \}_{i = 1}^n$, where $\boldsymbol{x}_{zi}$ is an image patch sampled from case $z$ and $y_{zi}$ is the ground truth class of its central pixel. We then denote $p_{zij}$ as the predicted probability of sample $\boldsymbol{x}_{zi}$ for the $j$'th class and its calibrated version as $\hat{p}_{zij}$. The estimated DSC can then be calculated using the predicted probabilities, in a fashion similar to the common soft DSC loss~\cite{mehrtash2020confidence, milletari2016v}. Specifically, $\text{DSC}_j$, which is real DSC of $\mathcal{D}^V$ for the $j$'th class, is estimated with:

\begin{equation}
    \text{sDSC}_j = \frac{1}{Z}\sum_{z=1}^{Z}\frac{2\sum_{i=1}^{n}\mathbbm{1}_{[\argmax_k p_{zik}=j]}\hat{p}_{zij}}{\sum_{i=1}^{n}\mathbbm{1}_{[\argmax_k p_{zik}=j]}+\sum_{i=1}^{n}\hat{p}_{zij}}.
    \label{eq:sDSC}
\end{equation}

For the CS TS, CS ATC and CS ATC-CS, we first calibrate $p_{zij}$ of the background class by matching class-wise average confidence and accuracy $\alpha^V_j$, then calibrate $p_{zij}$ of the foreground class $j$ by matching $\text{sDSC}_j$ and $\text{DSC}_j$, in order to adjust the model to fit the metric of DSC. We calibrate the background class according to accuracy instead of DSC because the $\text{sDSC}_j$ is always very high and cannot be easily decrease to match $\text{DSC}_j$, which we found makes optimization to match them meaningless. For CS DoC, as $d_j$ cannot be calculated with the difference between $\text{sDSC}_j$ and $\text{DSC}_j$, we keep $\hat{p}_{zij}=p_{zij}$ and estimate DSC on the target dataset $\mathcal{D}^{Te}$ on $j$'th class simply by $\text{DSC}_{CSDoCj}^{Te} = \text{DSC}_{j} - \text{sDSC}_j + \text{sDSC}_j^{Te}$.

\section{Experiments}

\begin{table}[t]
\centering
\caption{Evaluation on different tasks under varied types of domain shifts based on Mean Absolute Error (MAE). Lower MAE is better. Best results with lowest MAE in \textbf{bold}. Class-specific calibration as proposed (CS methods) improves all baselines. This is most profound in segmentation tasks, which present extreme class imbalance.}
\label{tabx1}
\newsavebox{\tableEstimation}
\begin{lrbox}{\tableEstimation}
\begin{tabular}{m{28mm}<{\centering}|m{18mm}<{\centering}|m{18mm}<{\centering}|m{18mm}<{\centering}||m{18mm}<{\centering}|m{18mm}<{\centering}|m{18mm}<{\centering}}
\hlineB{3}
Task & \multicolumn{3}{c||}{Classification} & 
\multicolumn{3}{c}{Segmentation}  \\
\hlineB{2}
Training dataset & CIFAR-10 & \multicolumn{2}{c||}{HAM10000} & \multicolumn{1}{c|}{ATLAS} & \multicolumn{2}{c}{Prostate} \\
Test domain shifts & Synthetic & Synthetic & Natural & Synthetic & Synthetic & Natural \\
\hlineB{2}
AC & 31.3 $\pm$ 8.2 & 12.3 $\pm$ 5.1 & 20.1 $\pm$ 13.4 & 35.6 $\pm$ 2.1 & 8.7 $\pm$ 4.9 & 18.7 $\pm$ 5.9 \\
\hline
QC~\cite{robinson2018real} & ----- & ----- & ----- & 3.0 $\pm$ 1.7 & 5.2 $\pm$ 6.6 & 19.3 $\pm$ 7.1 \\
\hline
TS~\cite{guo2017calibration} & 5.7 $\pm$ 5.6 & 3.9 $\pm$ 4.3 & 12.1 $\pm$ 8.3 & 9.7 $\pm$ 2.5 & 3.7 $\pm$ 5.4 & 9.2 $\pm$ 4.9 \\
VS~\cite{guo2017calibration} & 3.8 $\pm$ 2.1 & 4.2 $\pm$ 4.2 & 13.6 $\pm$ 9.6 & 11.4 $\pm$ 2.5 & 4.8 $\pm$ 5.1 & 11.2 $\pm$ 4.9 \\
NORCAL~\cite{pan2021model} & 7.6 $\pm$ 3.8 & 4.2 $\pm$ 4.6 & 13.7 $\pm$ 9.6 & 6.7 $\pm$ 2.4 & 5.8 $\pm$ 5.7 & 7.3 $\pm$ 4.7 \\
\textbf{CS TS} & 5.5$^\sim$ $\pm$ 5.6 & 3.7$^\sim$ $\pm$ 4.0 & 11.9$^\sim$ $\pm$ 8.0 & 1.6$^{**}$ $\pm$ 1.8 & 3.0$^{**}$ $\pm$ 5.7 & 7.8$^{**}$ $\pm$ 4.8 \\
\hline
DoC~\cite{guillory2021predicting} & 10.8 $\pm$ 8.2 & 4.6 $\pm$ 5.0 & 15.3 $\pm$ 9.7 & 4.2 $\pm$ 3.2 &  3.7 $\pm$ 5.8 & 13.9 $\pm$ 6.5  \\
\textbf{CS DoC} & 9.4$^{**}$ $\pm$ 7.2 & 4.5$^\sim$ $\pm$ 4.9 & 14.7$^*$ $\pm$ 9.2 & \textbf{1.3}$^{**}$ $\pm$ \textbf{1.9} & 3.5$^{*}$ $\pm$ 6.1 & 12.1$^{*}$ $\pm$ 5.9 \\
\hline
ATC~\cite{garg2022leveraging} & 4.6 $\pm$ 4.4 & 3.4 $\pm$ 3.9 & 7.1 $\pm$ 6.3 & 30.4 $\pm$ 1.8 & 8.6 $\pm$ 3.3 & 16.7 $\pm$ 5.3 \\
\textbf{CS ATC} & 2.8$^{**}$ $\pm$ 2.9 & \textbf{3.3}$^\sim$ $\pm$ \textbf{4.8} & \textbf{5.8}$^\sim$ $\pm$ \textbf{7.6} & 1.6$^{**}$ $\pm$ 1.5 & \textbf{1.1}$^{**}$ $\pm$ \textbf{1.7} & 4.3$^{**}$ $\pm$ 2.2 \\
\hline
TS-ATC~\cite{guo2017calibration, garg2022leveraging} & 5.3 $\pm$ 3.9 & 4.2 $\pm$ 4.2 & 7.3 $\pm$ 7.1 & 30.4 $\pm$ 1.8 & 8.5 $\pm$ 3.3 & 16.7 $\pm$ 5.3 \\
\textbf{CS TS-ATC }& \textbf{2.7}$^{**}$ $\pm$ \textbf{2.3} & 4.2$^\sim$ $\pm$ 5.4 & 5.9$^\sim$ $\pm$ 8.4 & \textbf{1.3}$^{**}$ $\pm$ \textbf{1.4} & 1.2$^{**}$ $\pm$ 1.7 & \textbf{4.2}$^{**}$ $\pm$ \textbf{2.2} \\
\hline
\end{tabular}
\end{lrbox}
\scalebox{0.75}{\usebox{\tableEstimation}}

{\raggedright \footnotesize{\quad $^*p$-value $<$ 0.05; $^{**}p$-value $<$ 0.01; $^\sim p$-value $	\geq$ 0.05 (compared with their class-agnostic counterparts)\par}}

\end{table}

\begin{figure}[t]
\centering
\includegraphics[width=0.75\textwidth]{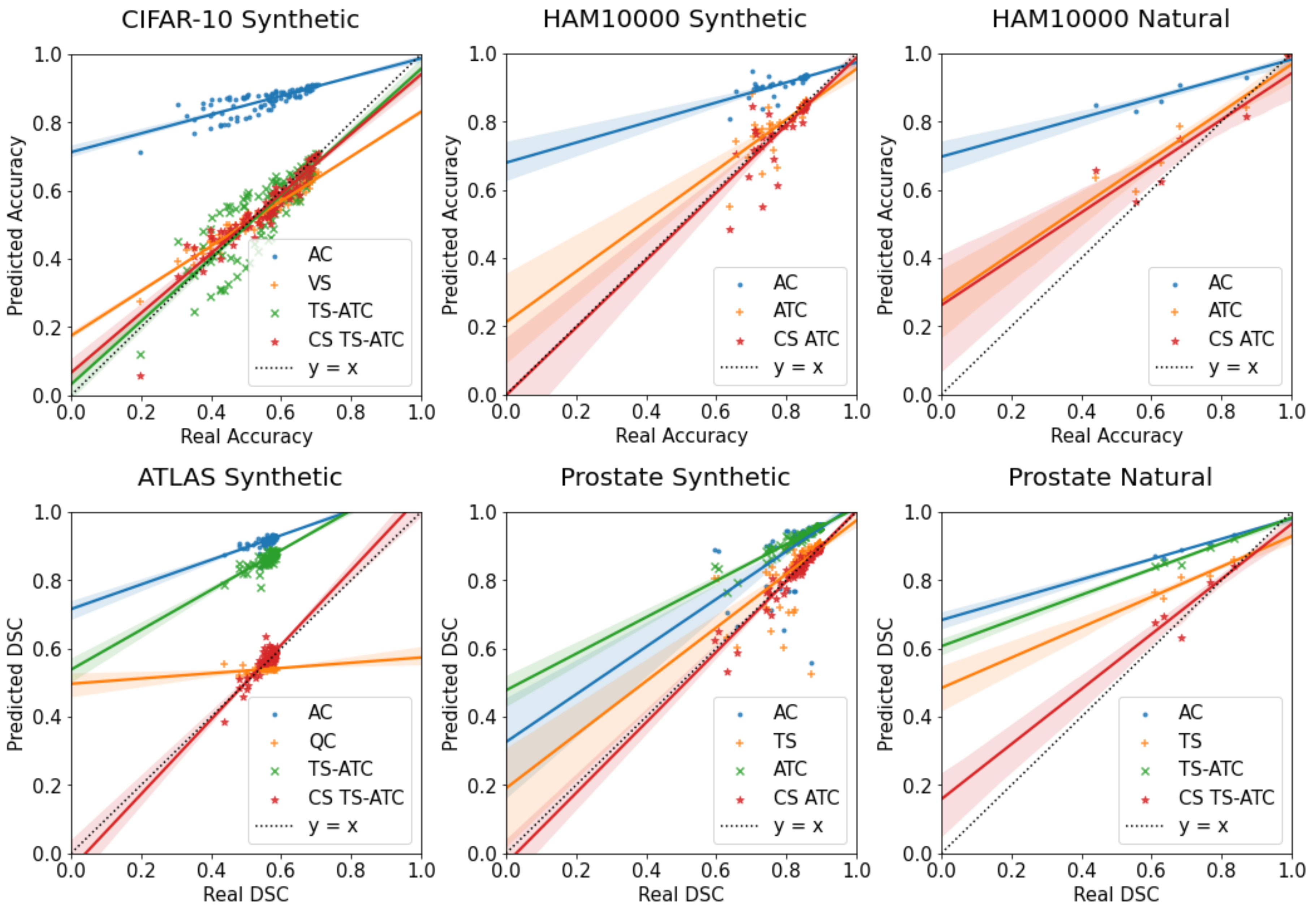}

\caption{Real versus predicted accuracy/DSC for classification/segmentation tasks respectively. Each point represents model performance on the unseen target data. We plot a baseline method (AC) in blue, the best among previous methods in orange, the best among the proposed methods in red and its class-agnostic counterpart in green. Best methods determined based on MAE.} \label{fig3}
\end{figure}

\textbf{Setting.} In our experiments, we first train a classifier $f$ with $\mathcal{D}^{Tr}$, then we make predictions on $\mathcal{D}^{V}$ and optimize the calibration parameters based on model output and the validation labels. Finally, we evaluate our model on $\mathcal{D}^{Te}$ and estimate the quality of the predicted results based on $\hat{\boldsymbol{p}}_i$. In addition to the confidence-based methods described in Sec.~\ref{sec:method}, for the segmentation tasks, we also compare with a neural network based quality control (QC) method~\cite{robinson2018real}. Following~\cite{robinson2018real}, we train a 3D ResNet-50 that takes as input the image and the predicted segmentation mask and predicts the DSC of the predicted versus the manual segmentation. For fair comparison, we train the QC model with the same validation data we use for model calibration.

\noindent\textbf{Datasets.} \textit{Imbalanced CIFAR-10:} We train a ResNet-32~\cite{he2016deep} with imbalanced CIFAR-10~\cite{krizhevsky2009learning}, using imbalance ratio of 100 following~\cite{cao2019learning}. We employ synthetic domain shifts using CIFAR-10-C~\cite{hendrycks2019robustness} that consists of 95 distinct corruptions. \textit{Skin lesion classification:} We train ResNet-50 for skin lesion classification with $c\!=\!7$ following~\cite{russakovsky2015imagenet, marrakchi2021fighting}. We randomly select 70\% data from HAM10000~\cite{tschandl2018ham10000} as the training source domain and leave the rest as validation. We also create synthetic domain shifts based on the validation data following~\cite{hendrycks2019robustness}, obtaining 38 synthetic test datasets with domain shifts. We additionally collect 6 publicly available skin lesion datasets which were collected in different institutions including BCN~\cite{combalia2019bcn20000}, VIE~\cite{rotemberg2021patient}, MSK~\cite{gutman2016skin}, UDA~\cite{gutman2016skin}, D7P~\cite{kawahara2018seven} and PH2~\cite{mendoncca2013ph} following ~\cite{yoon2019generalizable}. The detailed dataset statistics are summarized in supplementary material. \textit{Brain lesion segmentation:} We apply nnU-Net~\cite{isensee2021nnu} for all the segmentation tasks. We evaluate the proposed methods with brain lesion segmentation based on T1-weighted magnetic resonance (MR) images using data from Anatomical Tracings of Lesions After Stroke (ATLAS)~\cite{liew2018large} for which classes are highly imbalanced. We randomly select 145 cases for training and leave 75 cases for validation. We create synthetic domains shifts by applying 83 distinct transformations, including spatial, appearance and noise operations to the validation data. Details on implementation are in the supplementary material. \textit{Prostate segmentation:} We also evaluate on the task of cross-site prostate segmentation based on T2-weighted MR images following~\cite{liu2020saml}. We use 30 cases collected with 1.5T MRI machines from ~\cite{bloch2015nci} as the source domain. We randomly select 20 cases for training and 10 cases for validation. We create synthetic domains shifts using the same 83 transformations as for brain lesions. As target domain, we use the 5 datasets described in~\cite{lemaitre2015computer, bloch2015nci, litjens2014evaluation}, collected from different institutions using scanners of different field strength and manufacturers.

\noindent\textbf{Results and Discussion.} We calculate the Mean Absolute Error (MAE) and R2 Score between the predicted and the real performance (accuracy or DSC). We summarize MAE results in Table~\ref{tabx} and R2 Score results in supplementary material. We show scatter plots of predicted versus real performance in Fig.~\ref{fig3}. Each reported result is the average of two experiments with different random seeds. We observe in Table~\ref{tabx} that our proposed modifications for class-specific calibration improve all baselines. Improvements are especially profound in segmentation tasks, where class imbalance is very high. It might be because previous model calibration methods bias towards the background class significantly as a result of class imbalance. These results clearly demonstrate the importance of accounting for class imbalance in the design of performance-prediction methods, to counter biases induced by the data. Fig.~\ref{fig3} shows that the predicted performance by AC baseline is always higher than real performance because neural networks are commonly over-confident \cite{guo2017calibration}. QC is effective in predicting real DSC under synthetic domain shifts but under-performs with natural domain shifts.
Results in both Table~\ref{tabx} and Fig.~\ref{fig3} show that CS ATC (or CS TS-ATC) consistently performs best, outperforming prior methods by a large margin.

\section{Conclusion}
In this paper, we propose estimating model performance under domain shift for medical image classification and segmentation based on class-specific confidence scores. We derive class-specific variants for state-of-the-art methods to alleviate calibration bias incurred by class imbalance, which is profound in real-world medical imaging databases, and show very promising results. We expect the proposed methods to be useful for safe deployment of machine learning in real world settings, for example by facilitating decisions such as whether to deploy a given, pre-trained model based on whether estimated model performance on the target data of a deployment environment meets acceptance criteria.

\section*{Acknowledgements}
ZL is grateful for a China Scholarship Council (CSC) Imperial Scholarship. This project has received funding from the ERC under the EU's Horizon 2020 research and innovation programme (grant No 757173) and the UKRI London Medical Imaging \& Artificial Intelligence Centre for Value Based Healthcare, and a EPSRC Programme Grants (EP/P001009/1).


%
%
%
%
\bibliographystyle{splncs04}
\bibliography{reference}

\appendix

\section{Dataset visualization}

\begin{figure}
\centering
\includegraphics[width=\textwidth]{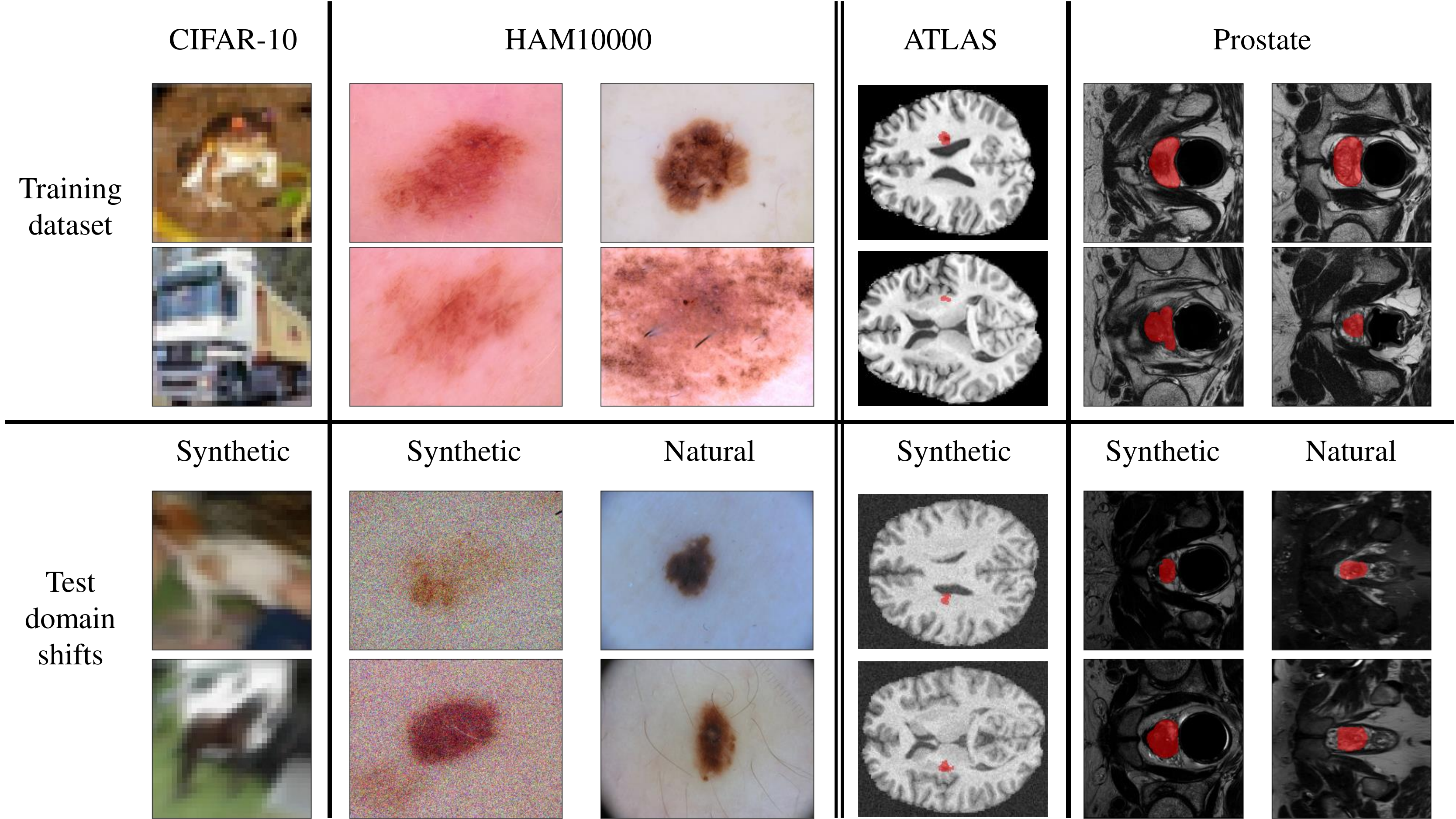}

\caption{Visualization of the datasets used in this study.} \label{figa1}
\end{figure}

\section{Dataset statistics of the skin lesion datasets}

\begin{table}
\centering
\caption{Dataset statistics of skin lesion classification datasets. }
\label{tabs}
\newsavebox{\tabledataset}
\begin{lrbox}{\tabledataset}
\begin{tabular}{m{10mm}<{\centering}|m{15mm}<{\centering}m{14mm}<{\centering}m{14mm}<{\centering}m{14mm}<{\centering}m{14mm}<{\centering}m{14mm}<{\centering}m{14mm}<{\centering}m{10mm}<{\centering}}
\hlineB{3}
 & nv (\%) & mel (\%) & bcc (\%) & df (\%) & bkl (\%) & vasc (\%) & akiec (\%) & total \\
\hlineB{2}
HAM & 6705 (67) & 1113 (11) & 514 (5) & 115 (1) & 1099 (11) & 142 (1) & 327 (3) & 10015 \\
BCN & 4206 (34) & 2857 (23) & 2809 (23) & 124 (1) & 1138 (9) & 111 (1) & 1168 (9) & 12413 \\
VIE & 4331 (99) & 34 (1) & 0 & 0 & 0 & 0 & 0 & 4365 \\
MSK & 2202 (62) & 826 (23) & 30 (1) & 5 ($<$1) & 470 (13) & 0 & 7 ($<$1) & 3540 \\
UDA & 408 (67) & 193 (31) & 3 ($<$1) & 2 ($<$1) & 7 (1) & 0 & 0 & 613 \\
D7P & 1150 (60) & 501 (26) & 84 (4) & 40 (2) & 90 (5) & 58 (3) & 0 & 1923 \\
PH2 & 160 (80) & 40 (20) & 0 & 0 & 0 & 0 & 0 & 200 \\
\hline
\end{tabular}
\end{lrbox}
\scalebox{0.95}{\usebox{\tabledataset}}
\end{table}

\section{Synthetic domain shifts for segmentation tasks}

\begin{table*}
\centering
\caption{List of all transformations that we use to generate synthetic domain shifts for segmentation. There are 24 types of operations with a total of 83 choices.}\label{tabteaug}
\newsavebox{\tableboxauglist}
\begin{lrbox}{\tableboxauglist}

\begin{tabular}{m{15mm}<{\centering}m{20mm}<{\centering}m{46mm}<{\centering}m{61mm}<{\centering}m{35mm}<{\centering}}
\hlineB{3}
Category & Operation ID & Operation Name & Description & Range of magnitudes \\
\hlineB{2}
\multirow{18}{*}{\tabincell{c}{Spatial \\ transfor- \\ mations}} 
& 1,2,3,4,5 & Scaling down & Scale down the image by factor 1 + $M$ & [0.05, 0.15, 0.25, 0.35, 0.45] \\
\cline{2-5}
& 6,7,8,9,10 & Scaling up & Scale up the image by factor 1 + $M$ & [0.05, 0.15, 0.25, 0.35, 0.45] \\
\cline{2-5}
& 11,12,13,14,15 & RotateFrontal ACW & Rotate the image along frontal axis anticlockwise & [5$\degree$, 15$\degree$, 25$\degree$, 90$\degree$, 180$\degree$] \\
\cline{2-5}
& 16,17,18,19 & RotateFrontal CCW & Rotate the image along frontal axis clockwise & [5$\degree$, 15$\degree$, 25$\degree$, 90$\degree$] \\
\cline{2-5}
& 20,21,22,23,24 & RotateSagittal ACW & Rotate the image along sagittal axis anticlockwise & [5$\degree$, 15$\degree$, 25$\degree$, 90$\degree$, 180$\degree$] \\
\cline{2-5}
& 25,26,27,28 & RotateSagittal CCW & Rotate the image along sagittal axis clockwise & [5$\degree$, 15$\degree$, 25$\degree$, 90$\degree$] \\
\cline{2-5}
& 29,30,31,32,33 & RotateLongitudinal ACW & Rotate the image along longitudinal axis anticlockwise & [5$\degree$, 15$\degree$, 25$\degree$, 90$\degree$, 180$\degree$] \\
\cline{2-5}
& 34,35,36,37 & RotateLongitudinal CCW & Rotate the image along longitudinal axis clockwise  & [5$\degree$, 15$\degree$, 25$\degree$, 90$\degree$] \\
\cline{2-5}
& 38,39,40,41 & Mirroring & Flip the sample in different planes & [Sagittal, Frontal, Axial, All] \\
\hline
\multirow{17}{*}{\tabincell{c}{Intensity \\ transfor- \\ mations}} & 42,43,44 & Gamma expansion & Scale $\boldsymbol{x}$ to [0,1], then $\mathcal{O}$($\boldsymbol{x}$) = $\boldsymbol{x}$$^{1 + \gamma}$, and scale it back & [0.1, 0.3, 0.5] \\
\cline{2-5}
& 45,46,47 & Gamma compression & Scale $\boldsymbol{x}$ to [0,1], then $\mathcal{O}$($\boldsymbol{x}$) = $\boldsymbol{x}$$^{1/(1 + \gamma)}$, and scale it back & [0.1, 0.3, 0.5] \\
\cline{2-5}
& 48,49,50 & Inverted gamma expansion & Do the gamma compression with the inverted image & [0.1, 0.3, 0.5] \\
\cline{2-5}
& 51,52,53 & Inverted gamma compression & Do the gamma expansion with the inverted image & [0.1, 0.3, 0.5] \\
\cline{2-5}
& 54,55,56 & Adding intensity & $\mathcal{O}$($\boldsymbol{x}$) = $\boldsymbol{x}$ + shift & [0.05, 0.15, 0.25] \\
\cline{2-5}
& 57,58,59 & Subtracting intensity & $\mathcal{O}$($\boldsymbol{x}$) = $\boldsymbol{x}$ - shift & [0.05, 0.15, 0.25] \\
\cline{2-5}
& 60,61,62 & Scaling up intensity histogram & $\mathcal{O}$($\boldsymbol{x}$) = $\boldsymbol{x}$ * (1 + scale) & [0.05, 0.15, 0.25] \\
\cline{2-5}
& 63,64,65 & Scaling down intensity histogram & $\mathcal{O}$($\boldsymbol{x}$) = $\boldsymbol{x}$ / (1 + scale) & [0.05, 0.15, 0.25] \\
\cline{2-5}
& 66,67,68 & Increasing contrast & Reduce the image mean, then $\mathcal{O}$($\boldsymbol{x}$) = $\boldsymbol{x}$ * (1 + scale), and add the mean value back & [0.05, 0.15, 0.25] \\
\cline{2-5}
& 69,70,71 & Decreasing contrast & Reduce the image mean, then $\mathcal{O}$($\boldsymbol{x}$) = $\boldsymbol{x}$ / (1 + scale), and add the mean value back & [0.05, 0.15, 0.25] \\
\hline
\multirow{7}{*}{\tabincell{c}{Noise \\ transfor- \\ mations}} & 72,73,74 & Blurring & Blur the image using Gaussian filter with different standard deviations & [0.5, 0.7, 0.9] \\
\cline{2-5}
& 75,76,77 & Sharpening & Sharpen the image using Laplacian of Gaussian filter with different standard deviations & [0.9, 0.7, 0.5] \\
\cline{2-5}
& 78,79,80 & Adding Gaussian noise & Add Gaussian noise with different standard deviations & [0.025, 0.075, 0.125] \\
\cline{2-5}
& 81,82,83 & Simulating low resolution & Scale down the image, and scale it back & [0.9, 0.7, 0.5] \\
\hline
\end{tabular}
\end{lrbox}
\scalebox{0.66}{\usebox{\tableboxauglist}}
\end{table*}

\section{Evaluation results based on R2 score}

\begin{table}
\centering
\caption{Evaluation of different tasks under varied types of domain shifts based on R2 Score. Higher R2 Score is better. The best results are in \textbf{bold}.}
\label{tabx}
\newsavebox{\tableEstimationR}
\begin{lrbox}{\tableEstimationR}
\begin{tabular}{m{28mm}<{\centering}|m{15mm}<{\centering}|m{14mm}<{\centering}|m{14mm}<{\centering}||m{14mm}<{\centering}|m{14mm}<{\centering}|m{14mm}<{\centering}}
\hlineB{3}
Task & \multicolumn{3}{c||}{Classification} & 
\multicolumn{3}{c}{Segmentation}  \\
\hlineB{2}
Training dataset & CIFAR-10 & \multicolumn{2}{c||}{HAM10000} & \multicolumn{1}{c|}{ATLAS} & \multicolumn{2}{c}{Prostate} \\
Test domain shifts & Synthetic & Synthetic & Natural & Synthetic & Synthetic & Natural \\
\hlineB{2}
AC & -7.88 & -3.15 & -0.70 & -157.94 & -1.29 & -4.52 \\
\hline
QC & ----- & ----- & ----- & -0.50 & -0.63 & -5.02 \\
\hline
TS & 0.46 & 0.22 & 0.38 & -11.59 & 0.02 & -0.56 \\
VS & 0.84 & 0.18 & 0.19 & -16.14 & -0.12 & -1.16 \\
NORCAL & 0.39 & 0.11 & 0.19 & -5.26 & -0.51 & -0.07 \\
\textbf{CS TS} & 0.48 & 0.30 & 0.40 & 0.28 & 0.04 & -0.19 \\
\hline
DOC & -0.55 & -0.07 & 0.04 & -2.47 & -0.08 & -2.38 \\
\textbf{CS DOC} & -0.19 & -0.04 & 0.13 & 0.30 & -0.13 & -1.59 \\
\hline
ATC & 0.66 & \textbf{0.37} & \textbf{0.74} & -114.97 & -0.93 & -3.39 \\
\textbf{CS ATC} & 0.86 & 0.20 & 0.73 & 0.40 & \textbf{0.91} & 0.67 \\
\hline
TS-ATC & 0.63 & 0.17 & 0.70 & -114.80 & -0.92 & -3.37 \\
\textbf{CS TS-ATC} & \textbf{0.89} & -0.10 & 0.70 & \textbf{0.54} & 0.90 & \textbf{0.68} \\
\hline
\end{tabular}
\end{lrbox}
\scalebox{0.8}{\usebox{\tableEstimationR}}
\end{table}

\section{Boxplots of MAE results}
\begin{figure}
\centering
\includegraphics[width=\textwidth]{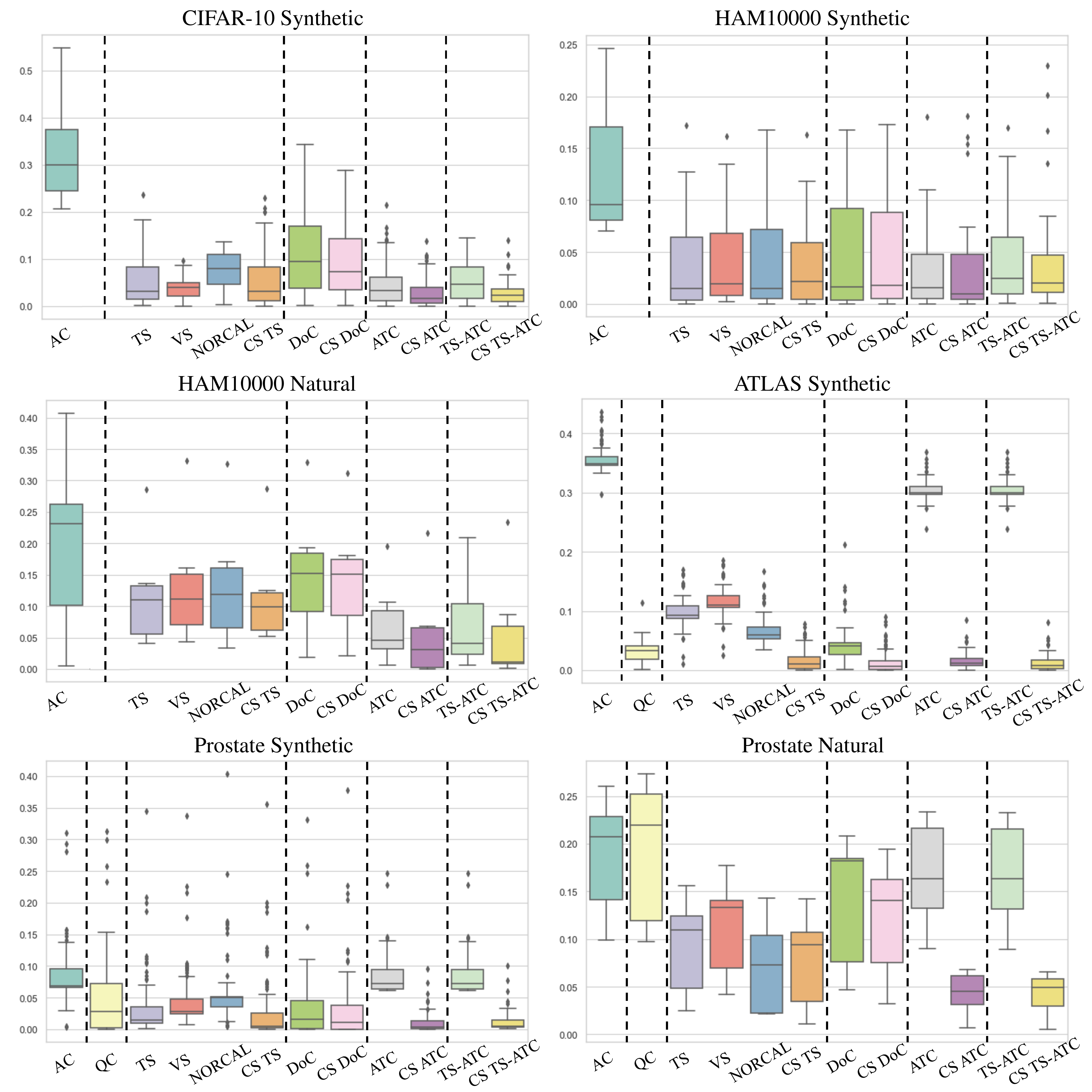}

\caption{Boxplots of the MAE results for different methods. The methods with the same original formulation are grouped together.} \label{figa2}
\end{figure}

\end{document}